\documentclass[sigconf]{acmart}
\AtBeginDocument{%
  }

\setcopyright{acmlicensed}
\copyrightyear{2025}
\acmYear{2025}
\setcopyright{acmlicensed}
\acmConference[MM '25] {Proceedings of the 33rd ACM International Conference on Multimedia}{October 27--31, 2025}{Dublin, Ireland.}
\acmBooktitle{Proceedings of the 33rd ACM International Conference on Multimedia (MM '25), October 27--31, 2025, Dublin, Ireland}
\acmISBN{979-8-4007-2035-2/2025/10}
\acmDOI{10.1145/XXXXXX.XXXXXX}

\usepackage{bbding}
\usepackage{microtype}
\usepackage{graphicx}
\usepackage{tabularx}

\usepackage{times}
\usepackage{latexsym}
\usepackage{subcaption}
\usepackage{amsmath}
\usepackage{times}
\usepackage{latexsym}
\usepackage{multirow}
\begin{document}

\title[Impact of Stickers on Multimodal Sentiment and Intent in
Social Media]{Impact of Stickers on Multimodal Sentiment and Intent in
Social Media: A New Task, Dataset and Baseline}

\author{Yuanchen Shi}
\orcid{0009-0001-5651-5084}
\affiliation{%
  \institution{School of Computer Science and Technology,}
  \department{Soochow University,}
  \city{Suzhou}
  \country{China}}
\email{20227927002@stu.suda.edu.cn}

\author{Biao Ma}
\affiliation{%
  \institution{School of Computer Science and Technology,}
  \department{Soochow University,}
  \city{Suzhou}
  \country{China}}
\email{biaoma@alu.suda.edu.cn}

\author{Longyin Zhang}
\orcid{0000-0002-0542-6508}
\affiliation{%
  \institution{Institute for Infocomm Research, A*STAR,}
  \department{Aural \& Language Intelligence,}
  \city{Singapore}
  \country{Singapore}}
\email{zhang\_longyin@i2r.a-star.edu.sg}

\author{Fang Kong\textsuperscript{\textasteriskcentered}}
\orcid{0000-0002-7102-0143}
\affiliation{%
  \institution{School of Computer Science and Technology,}
  \department{Soochow University,}
  \city{Suzhou}
  \country{China}
}
\email{kongfang@suda.edu.cn}

\thanks{\textsuperscript{\textasteriskcentered} Corresponding author}

\renewcommand{\shortauthors}{Shi et al.}

\begin{abstract}
Stickers are increasingly used in social media to express sentiment and intent. Despite their significant impact on sentiment analysis and intent recognition, little research has been conducted in this area. To address this gap, we propose a new task: \textbf{M}ultimodal chat \textbf{S}entiment \textbf{A}nalysis and \textbf{I}ntent \textbf{R}ecognition involving \textbf{S}tickers (MSAIRS). Additionally, we introduce a novel multimodal dataset containing Chinese chat records and stickers excerpted from several mainstream social media platforms. Our dataset includes paired data with the same text but different stickers, the same sticker but different contexts, and various stickers consisting of the same images with different texts, allowing us to better understand the impact of stickers on chat sentiment and intent. We also propose an effective multimodal joint model, MMSAIR, featuring differential vector construction and cascaded attention mechanisms for enhanced multimodal fusion. Our experiments demonstrate the necessity and effectiveness of jointly modeling sentiment and intent, as they mutually reinforce each other's recognition accuracy. MMSAIR significantly outperforms traditional models and advanced MLLMs, demonstrating the challenge and uniqueness of sticker interpretation in social media. Our dataset and code are available on \url{https://github.com/FakerBoom/MSAIRS-Dataset}.
\end{abstract}



\maketitle

\section{Introduction}

With the proliferation of social media platforms, users increasingly employ these channels as primary vehicles for expressing emotional states \cite{gaind2019emotion} and communicative intentions \cite{purohit2015intent}. Sentiment analysis aims to classify content as positive, negative, or neutral, while intent recognition categorizes the underlying communicative purpose. Sentiment and intent typically co-occur in discourse, with emotional states often driving specific intentions, while particular intentions frequently reveal underlying affective dispositions \cite{lewis2005getting}.

\begin{figure}[!t]
\begin{center}
\includegraphics[scale=0.45]{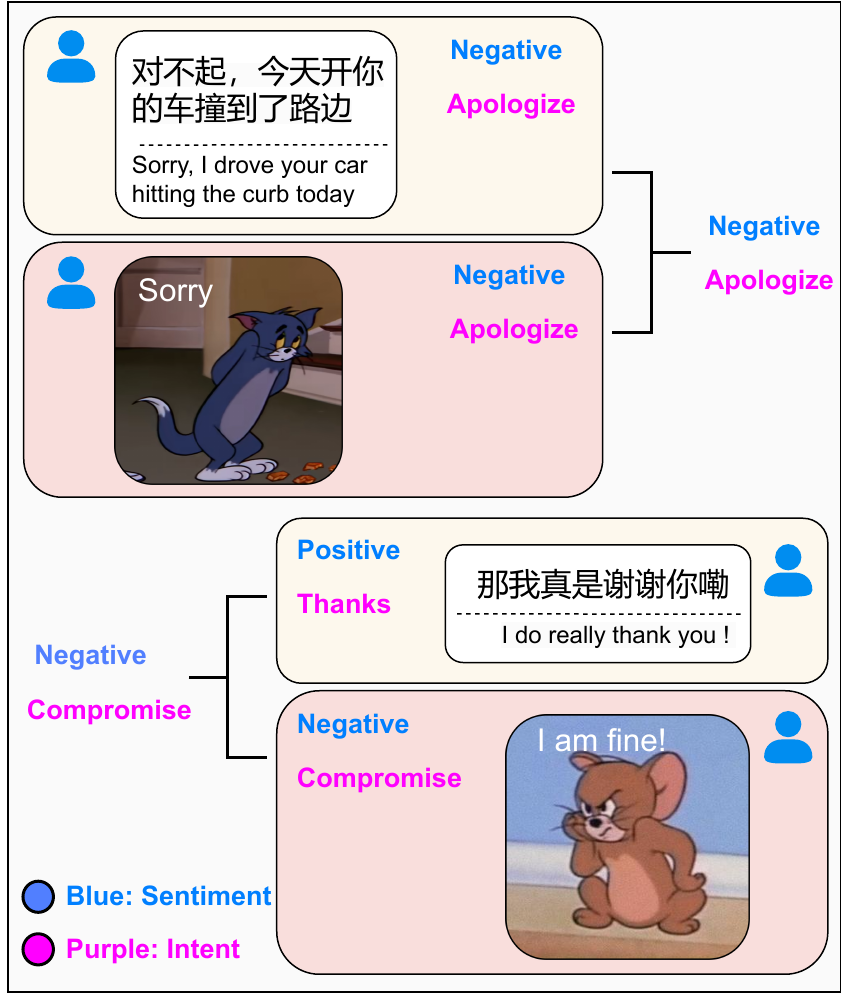} 
\caption{A chat record from a social media platform. Only by combining stickers can we discern the true pessimism and complaint the second man wants to express.}
\label{fig.1}
\end{center}
\end{figure}

In chatting applications, social platforms, and media comments, a plethora of images, commonly referred to as stickers, emoticons, emojis or memes\footnote{In this paper, we collectively refer to them as "stickers".}, can be observed. These images serve as a substitute for expressing thoughts that are challenging to convey by text alone, aiding individuals in better expressing sentiment and intent \cite{ge-etal-2022-towards}. However, this field hasn't been extensively researched due to issues such as text-image misalignment and lack of suitable datasets. 

Currently, numerous studies have separately investigated multimodal sentiment analysis \cite{abdullah2021multimodal} and intent recognition \cite{huang2023effective}. However, a handful have combined these two tasks. Most studies explore the fusion of modalities like real photos and text \cite{yang2019exploring}, video and text \cite{seo2022end}, audio and video with text \cite{akbari2021vatt}, etc., with limited research focusing on stickers and chat text. In social media, people prefer using stickers to express themselves. Stickers are often more convenient compared to text, allowing for a vivid and direct expression of ideas \cite{2022An}. As shown in Figure ~\ref{fig.1}\footnote{English translation is below the dotted line, as is the case with the other Figures.}, the text and sticker send by the first man both convey a negative sentiment and an intent to apologize, making it easy to comprehend his overall message. In contrast, while the text send by the second man indicates optimism and gratitude, the sticker shows a sense of pessimism and resignation, implying that he desires to express negativity and complaints. In such situations, when it might be difficult to express directly through language, a sticker can easily convey inner feelings. Thus, only by considering stickers simultaneously can we accurately determine sentiment and intent. In addition, it can be seen that sentiment and intent are interrelated. Although the context seems to express gratitude, it is clear that the intent cannot be gratitude after receiving a negative sentiment, so it must be a compromise. Similarly, based on the intent of compromise, it can be seen that the sentiment is definitely negative. Therefore, sentiment and intent need to be handled together. Consequently, we introduce Multimodal chat Sentiment Analysis and Intent Recognition involving Stickers (MSAIRS), a completely new task, as well as a dataset of the same name to support our research.

MSAIRS task is challenging due to the abstract nature of stickers. Stickers are often multimodal themselves, containing both image and text\footnote{We refer to the chat text as "context", and the text within stickers as "sticker-text".}, leading to variations when the image remains the same but sticker-text differs. Therefore, the task requires adept handling of context, stickers, and sticker-texts, demanding valid multimodal fusion methods. To address these challenges, we introduce a simple yet effective baseline: a joint Model for Multimodal Sentiment Analysis and Intent Recognition (MMSAIR). MMSAIR separately processes the input context, sticker and sticker-text, and integrates these multimodal components through a sophisticated fusion approach featuring cascaded multi-head attention mechanisms, differential vector construction, and feature concatenation, ultimately enabling accurate joint prediction of sentiment and intent in social media conversations. Our experiments demonstrate that sentiment analysis and intent recognition mutually reinforce each other, as proven by the improved performance when these tasks are modeled jointly rather than separately. Experimental results show that compared with many pre-trained models and multimodal large language models (MLLMs), our model performs significantly better, indicating the necessity of simultaneously integrating textual context and visual sticker information for comprehensive understanding of social media communications.

The contributions of this paper are as follows:

\begin{itemize} 
\item We introduce MSAIRS task and dataset to investigate the impact of stickers on multimodal chat sentiment and intent. 
\item We introduce a novel multimodal baseline, MMSAIR, for the joint task of multimodal sentiment analysis and intent recognition. Experiments show that MMSAIR performs better than other mainstream models. 
\item We validate the necessity and benefits of jointly studying sentiment and intent in sticker-based communications, showing that these aspects are intrinsically connected. 
\item To the best of our knowledge, we are the first to investigate the joint task of multimodal sentiment analysis and intent recognition involving stickers. 
\end{itemize}

\begin{figure*}[!t]
\begin{center}
\includegraphics[scale=0.5]{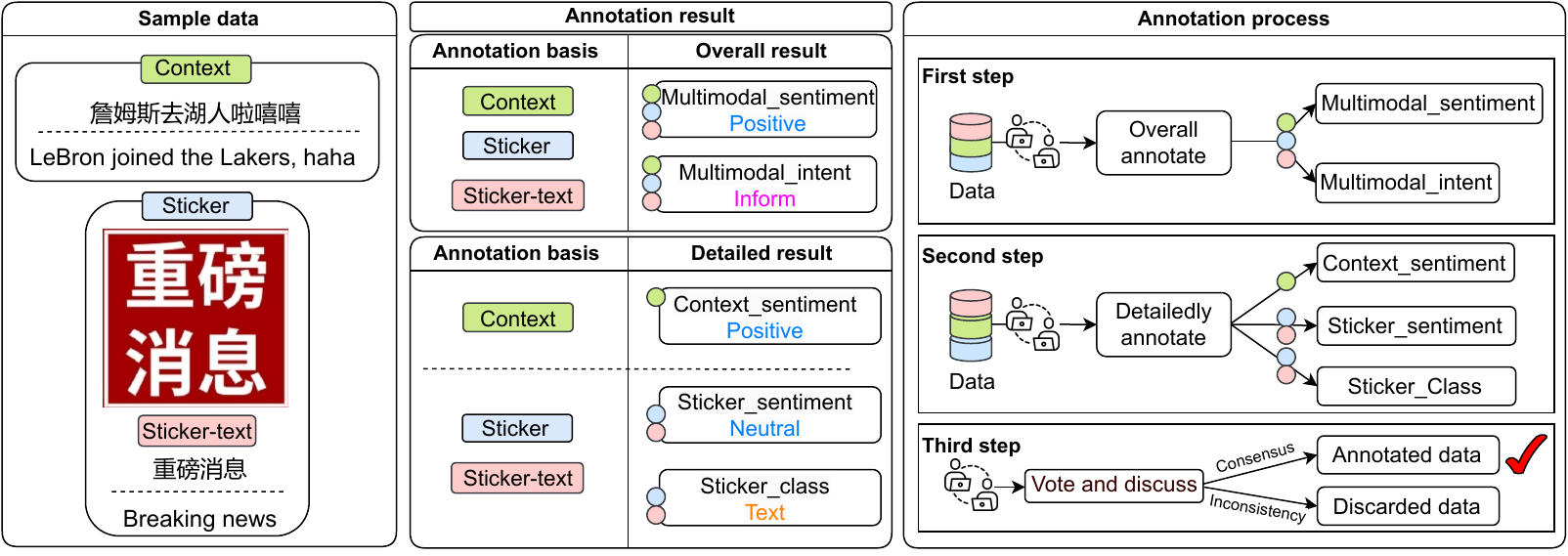} 
\caption{Annotating process and the annotation results obtained using this process for a sample from our dataset.}
\label{fig.2}
\end{center}
\end{figure*}

\section{ Related Work }

\subsection{Multimodal Sentiment and Intent Researches}

Multimodal sentiment analysis and intent recognition have been extensively studied in recent years. Existing studies typically focus on combining multiple modalities such as text, images, audio, and video to enhance the performance of sentiment and intent understanding. For instance, \cite{yu-etal-2020-ch} introduced CH-SIMS, a Chinese multimodal sentiment analysis dataset with fine-grained annotations across text, audio, and video modalities. \cite{mao-etal-2022-sena} presented M-SENA, an integrated platform for multimodal sentiment and emotion analysis that effectively combines visual, acoustic, and textual modalities to capture affective information. In the intent recognition domain, \cite{zhang2024mintrec20largescalebenchmarkdataset} proposed MIntRec 2.0, a large-scale benchmark dataset for multimodal intent recognition and out-of-scope detection in conversations, covering text, audio, and visual modalities. Furthermore, recent works have explored joint modeling of emotions and intents in multimodal conversations. \cite{liu2024emotionintentjointunderstanding} presented a benchmarking dataset for joint understanding of emotion and intent in multimodal conversations. Similarly, \cite{9961847} proposed EmoInt-Trans, a multimodal transformer model designed specifically for identifying emotions and intents in social conversations, along with a large-scale multi Emotion and Intent guided Multimodal Dialogue (EmoInt-MD) dataset.

However, existing multimodal datasets and methods rarely consider stickers, which are prevalent in social media communication and significantly influence sentiment and intent expression.

\subsection{Sticker-related Studies}

The rise of internet stickers has spurred numerous studies \cite{shifman2013memes, tang2019emoticon}. Sociologically, stickers are seen as cultural symbols, representing a key aspect of global internet culture \cite{2021Do, zhao2023sticker820kempoweringinteractiveretrieval}. In recent years, several tasks and datasets involving stickers have been proposed, such as sticker-based dialogue summarization \cite{shi2024integrating}, sticker empathetic response generation \cite{zhang-etal-2024-stickerconv}, and sticker retrieval \cite{chen2024deconfounded}.

Stickers are popular due to their rich metaphorical content, which reflects users' sentiments and intents. \cite{french2017image} explored the sentimental correlation between stickers' implicit semantics and social media discussions, highlighting their role in sentiment analysis. \cite{prakash2021hybrid} used neural networks for facial recognition in memes for sentiment analysis. However, not all stickers depict human portraits. \cite{pranesh2020memesem} applied transfer learning for sentiment analysis on stickers with varied styles, conducting unimodal and bimodal analysis. Specific emotions in stickers, like hatred and humor, have also been studied. \cite{lestari2019irony} analyzed the irony in stickers linguistically, while \cite{tanaka2022learning} created a humor analysis dataset, asserting that humor stems from incongruity between stickers and captions. \cite{qu2023evolution} examined hateful emotions in stickers, showing the real-world impact of such content.

Stickers inherently carry intent, with sentiments amplifying these intentions \cite{saha2021emotion}. To explore sticker intent on social media, \cite{jia2021intentonomy} introduced a dataset for recognizing intent behind social media images. \cite{10.1145/3477495.3532019} presented a comprehensive sticker dataset with labels like subjects, metaphors, aggressiveness, and emotions. However, current intent recognition often focuses on unimodal information, neglecting the link between sentiment and intent. To address these gaps, we propose a multimodal sentiment analysis and intent recognition dataset tailored for social media stickers.

\section{MSAIRS Dataset}

\subsection{Data Preparation}

To study the sentiment and intent in multimodal chat conversations with stickers, we introduce the MSAIRS dataset. Referring to the CSMSA dataset \cite{ge-etal-2022-towards}, MSAIRS retains the sentiment labels while adding multimodal intent labels. Our research team manually collect over 5k chat records or comments with clear intent and stickers from social media platforms such as WeChat, TikTok, and QQ. We choose manual collection instead of automatic generation because current AI models can primarily generate formal or realistic images but lack the capability to create abstract stickers with ironic or humorous textual content. Moreover, social media conversations often contain real-time internet slang, metaphors, and cultural references, which AI-generated content cannot authentically replicate.

All collected posts and chat records are publicly accessible, complying with the respective social media platforms' policies. To respect user privacy, we anonymize the data by replacing real usernames with third-person pronouns or generic placeholders. For each data entry, we ensure that both the context and sticker are sent by the same individual. Additionally, for stickers containing text, we utilize PaddleOCR \cite{du2020pp} to automatically extract the sticker-text and then incorporate it into our dataset.

\begin{table*}[ht] 
\centering   
\caption{\label{tbintent}
The quantity and proportion of each intent label with its brief description in dataset.}
\begin{tabular}{l|l|c|c}   
\toprule
       \textbf{Intent} & \centering \textbf{Brief description} & \textbf{Quantity} & \textbf{Proportion} \\
\midrule    Comfort &Relax someone by making them feel at ease.  &103 &3.30\% \\  \hline
 Oppose   & Disagree with or hinder someone or something. & 173&5.55\% \\  \hline
Greet  & Meet or welcome someone with a kind wave, smile or word. & 106& 3.40\%\\  \hline
Complain  & Express discontent or annoyance to someone or something. & 278& 8.92\%\\  \hline
Ask for help   &  Express the need for someone's help or guidance.& 166& 5.32\%\\  \hline
Taunt    & Use irony or sarcasm to mock someone. & 158&5.07\% \\  \hline
 Apologize   & Regret a mistake and request forgiveness. & \textbf{58}&\textbf{1.86\%} \\  \hline
Introduce  &Detailedly describe or recommend someone or something. & 199 & 6.38\%   \\  \hline
 Guess  &Speculate on the reasons or about the outcome.  & 106&3.40\% \\  \hline
Advise & Propose something or suggest doing something. & 179&5.74\% \\  \hline
 Compromise &Reluctantly make a concession or acceptance out of necessity.  &150 &4.81\% \\  \hline
 Praise &Admire or speak highly of someone or something.  &181 &5.81\% \\  \hline
 Inform & Let someone know about something. & 221& 7.09\%\\  \hline
Flaunt & Show off exaggeratedly in order to gain attention. &115 &3.69\% \\  \hline
Criticize & Censure or comment sharply on someone or something. & 124& 3.98\%\\  \hline
Thank & Express gratitude for someone's help or kindness. & 73&2.34\% \\  \hline
 Agree & Concur on something or with someone's viewpoint. & 143&4.59\% \\  \hline
Leave & Get away temporarily possibly to conclude the conversation. &109 &3.50\% \\  \hline
Query & Inquire others in order to find out something. & \textbf{311}&\textbf{9.97\%} \\  \hline
Joke & Make exaggerated or humorous statements for entertainment. & 165&5.29\% \\  \midrule
Overall & Sum of all intent labels. & \textbf{3118}&\textbf{100\%} \\  \bottomrule
\end{tabular}   

\end{table*}

\subsection{Data Annotation}

We employ five linguistics professionals, each with extensive experience in annotating Chinese datasets. The detailed annotation process is shown in Figure~\ref{fig.2}. Each annotator is required to label five categories. The \textit{context\_sentiment} and \textit{sticker\_sentiment} labels separately analyze the sentiment of the context and sticker, while the \textit{multimodal\_sentiment} and \textit{multimodal\_intent} labels represent the overall sentiment and intent considering both modalities. 

For intent labels, we refer to Mintrec \cite{zhang2022mintrec} but replace several labels due to differences in application scenarios. Mintrec originates from television dramas; thus, some labels, such as "Prevent", typically require concrete actions in real-world scenarios to stop someone or something, making them unsuitable for social media contexts. We remove such labels and introduce labels more common in social media, such as "Compromise", to capture users' expressions of helplessness or resignation. Finally, we categorize intents into twenty classes as listed in Table~\ref{tbintent}.

In Figure~\ref{fig.2}, the context alone might indicate informing, flaunting, or taunting intents. Only by simultaneously considering the sticker can we accurately determine that the intent is informing. Due to such ambiguity, annotating intent based solely on a single modality often leads to multiple plausible labels. Therefore, we assign intent labels exclusively at the multimodal level. Additionally, the \textit{sticker\_class} label categorizes stickers into four broad classes: real person, real animal, virtual entity (e.g., cartoon), and text-only. We further subdivide the first three classes based on whether there is text in the image, ultimately obtaining the sticker style distribution shown in Table \ref{style}. This classification highlights the diverse styles and preferences of social media stickers and supports further research.

To ensure data accuracy and credibility, each entry is retained only if three or more professionals annotate all the same labels. Otherwise, the entry is discarded. For entries where annotators do not agree, we engage in discussions to strive for consensus. If consensus cannot be reached, the data is discarded. After manual review, we retain 3.1k pieces with consistent labels.

\begin{table}[t]   \footnotesize
\caption{\label{tbcom}
Comparison of several multimodal datasets. \textit{t,v,a,i} represent text, video, audio, and image, respectively. \textit{SA} and \textit{IR} stand for sentiment analysis and intent recognition.
}
\centering   
\begin{tabular}{l|ccccc}   
\toprule
   \textbf{Dataset} & \textbf{Size} & \textbf{Modality} & \textbf{SA} &\textbf{IR}\\
\midrule

 MDID\cite{kruk2019integrating}&1299 &t,v & \XSolid& \Checkmark\\
 MIntRec\cite{zhang2022mintrec}  & 2224& t,v,a& \XSolid& \Checkmark\\
 CH-SIMS\cite{yu-etal-2020-ch} & 2281 & t,v,a&\Checkmark & \XSolid\\
   CSMAS\cite{ge-etal-2022-towards}&1564  &t,i &\Checkmark & \XSolid\\
MSAIRS(ours)&3118&t,i&\Checkmark&\Checkmark\\
\bottomrule  
\end{tabular}   

\end{table}

\subsection{Manual Annotation Necessity and AI Limitations}

We also use the GPT-4o model to categorize the overall sentiment and intent of the text and sticker. While GPT-4o performs well in single-modality sentiment annotation and achieves high consistency with human experts, its performance in multimodal sentiment and intent annotation is significantly less reliable. For example, as shown in Figure~\ref{ss}, the text “Getting out, pls” combined with the sticker is interpreted by GPT-4o as expressing a positive sentiment and an advising intent. However, from a human perspective, it is clear that the combination conveys a sarcastic tone, blaming someone in a passive-aggressive manner. Such discrepancies highlight the limitations of GPT-4o in understanding the nuanced interplay between text and stickers. These issues are further reflected in the experimental results in \ref{exre}, where GPT-4o struggles to align with human annotations in multimodal tasks. Therefore, we conclude that manual annotation is essential for ensuring the accuracy and reliability of the dataset.

\begin{table}[!t] \footnotesize
\centering   
\caption{\label{tb1}
Statistics on the quantity and proportion of the sentiment labels of different modalities in dataset.
}
\begin{tabular}{l|l|c|c}   
\toprule
   \textbf{Modality} & \textbf{Sentiment} & \textbf{Quantity} & \textbf{Proportion}\\
\midrule   &Positive  & 728     & 23.35\%\\
   Context&Negative & 921 &  29.54\%\\ 
   &Neutral & 1469 &  47.11\%\\
\midrule   &Positive  & 1115     & 35.76\%\\
   Sticker&Negative & 1166 &  37.40\%\\ 
  &Neutral & 837 &  26.84\%\\
\midrule   &Positive  & 1083     & 34.74\%\\
   Multimodal&Negative & 1358 &  43.55\%\\ 
   &Neutral & 677 &  21.71\%\\
\bottomrule  
\end{tabular}   
\end{table}

\subsection{Data statistics}

\begin{figure}[t]
\begin{center}
\includegraphics[width=0.4\textwidth,  keepaspectratio]{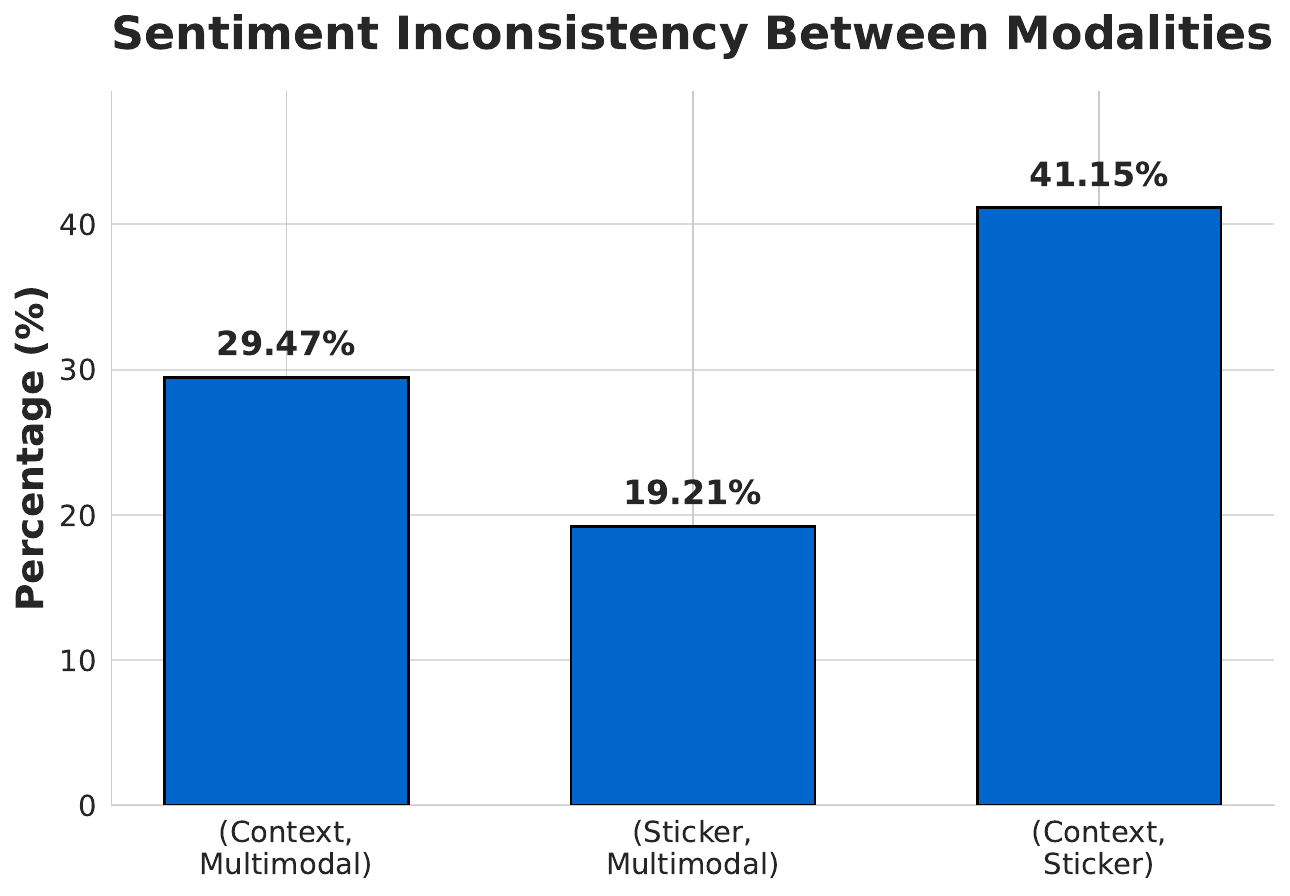} 
\caption{Inconsistency in sentiment labels across modalities.}
\label{incon}
\end{center}
\end{figure}

MSAIRS comprises 3.1k instances of data containing both context and stickers. In Table \ref{tbcom}, we list the comparison between MSAIRS and several mainstream multimodal sentiment or intent datasets. As can be seen, MSAIRS is the first dataset to include both sentiment analysis and intent recognition tasks. The quantity and proportion of multimodal intent labels is shown in Table ~\ref{tbintent}, which closely aligns with the actual proportions found on social media. Table ~\ref{tb1} presents the distribution of sentiment labels. We find that there are many samples where the sentiment labels obtained from different modalities are not consistent, which is demonstrated specifically in Figure ~\ref{incon}. This indicates that relying solely on the context or sticker may not accurately determine the overall sentiment, emphasizing the importance of multimodal holistic analysis.
Analysis of sticker styles in Table \ref{style} reveals notable user preferences in social media communication. Cartoon-style stickers strongly dominate the dataset, suggesting users favor the exaggerated expressiveness that virtual characters can convey compared to realistic imagery. Additionally, approximately 52.02\% of all stickers incorporate textual elements, highlighting the prevalent multimodal nature of stickers where visual content is often enhanced with text.

Our dataset also includes 4\% instances where the same context paired with different stickers results in varying sentiment and intent labels, as well as 31\% examples where identical stickers combined with different contexts yield distinct classifications. Both scenarios are described in detail in Section \ref{illex}. These complementary examples demonstrate the bidirectional influence between context and stickers - contexts can alter a sticker's interpretation while stickers can dramatically shift a context's perceived sentiment and intent. This interplay illustrates the indispensable role of stickers in sentiment analysis and intent recognition on social media, underscoring the necessity of our research.

\begin{table}[t]   \footnotesize
\caption{\label{style}
\textit{P}, \textit{A}, \textit{C} represents People, Animal, and Cartoon, without texts in the sticker. \textit{-t} represents that the sticker image contains texts. \textit{Text} means there is only text in the sticker.
}
\centering   
\begin{tabular}{l|ccccccc}   
\toprule
  & \textbf{P} & \textbf{A} & \textbf{C} & \textbf{P-t} & \textbf{A-t} & \textbf{C-t} & \textbf{Text}\\
\midrule
Quantity&64 &100 &1301 &118 &197 &\textbf{1307} &31\\
Proportion(\%)& 2.05& 3.21& 41.73& 3.78& 6.32& \textbf{41.92}& 0.99\\

\bottomrule   
\end{tabular}   
\end{table}

\subsection{Illustrative examples}
\label{illex}

\begin{figure}[!t]
\begin{center}
\includegraphics[width=0.35\textwidth,  keepaspectratio]{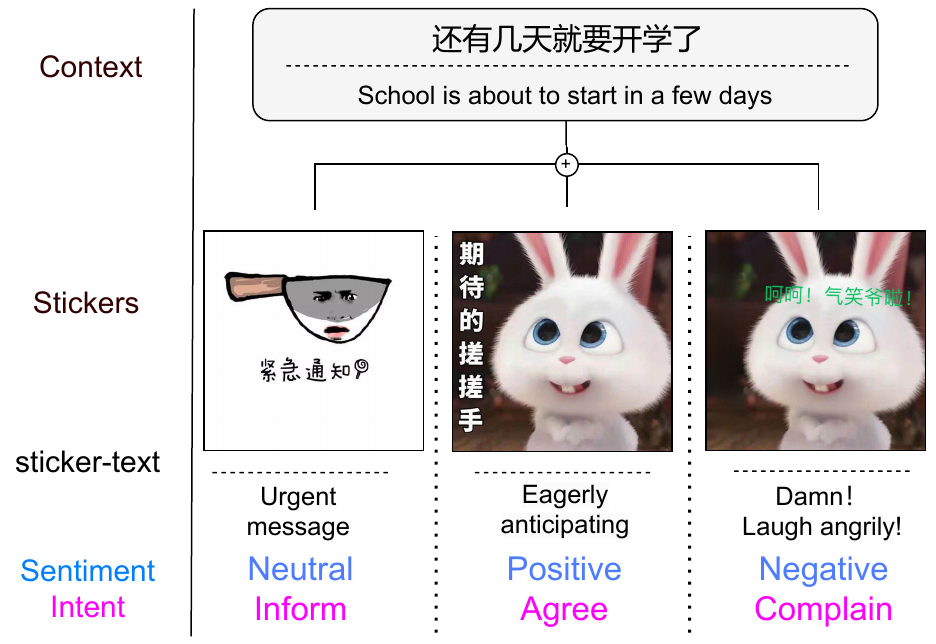} 
\caption{Examples illustrating how the same context, accompanied by different stickers or sticker-texts, can convey entirely distinct sentiment and intent.}
\label{fig.4}
\end{center}
\end{figure}

\begin{figure}[!t]
\begin{center}
\includegraphics[width=0.35\textwidth,  keepaspectratio]{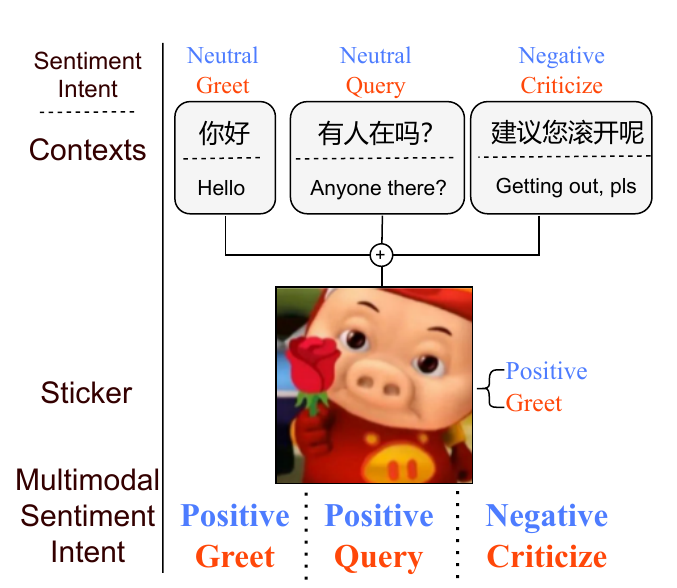} 
\caption{Examples illustrating how the same sticker with different contexts can convey entirely distinct sentiment and intent.}
\label{ss}
\end{center}
\end{figure}

Including the study of stickers can significantly contribute to the research on chat sentiment and intent. In figure ~\ref{fig.4}, the context "School is about to start in a few days" conveys a neutral sentiment with several possible intents such as inform and complain. The first sticker which expresses a neutral sentiment aims to inform others urgently. The rabbit in the second sticker depicts a joyful expression, signifying a positive sentiment and approval of the start of the school. Despite the visual similarity between the second and third sticker, the sticker-text "Damn! Laugh angrily!" reveals strong resistance and complaint about the start of school, forcing a smile negatively. Similarly, figure \ref{ss} demonstrates how the same sticker can convey drastically different sentiments and intents when paired with different contexts. As illustrated, the neutral greeting sticker featuring a cartoon character transforms significantly across three scenarios: maintaining its positive greeting intent with a friendly "Hello" context, shifting to a neutral query when asking "Anyone there?", and conveying negative criticism when paired with the dismissive text "Getting out, pls.". 

From these examples, we can clearly see the contradiction between unimodal and multimodal sentiments and intents. Sometimes the context plays a decisive role and sometimes it depends on the sticker, which is elusive as the context and sticker change. As a result, in order to recognize sentiment and intent in social media conversations, context, stickers, and sticker-text must be considered holistically.

\begin{figure*}[!ht]
\begin{center}
\includegraphics[scale=0.6]{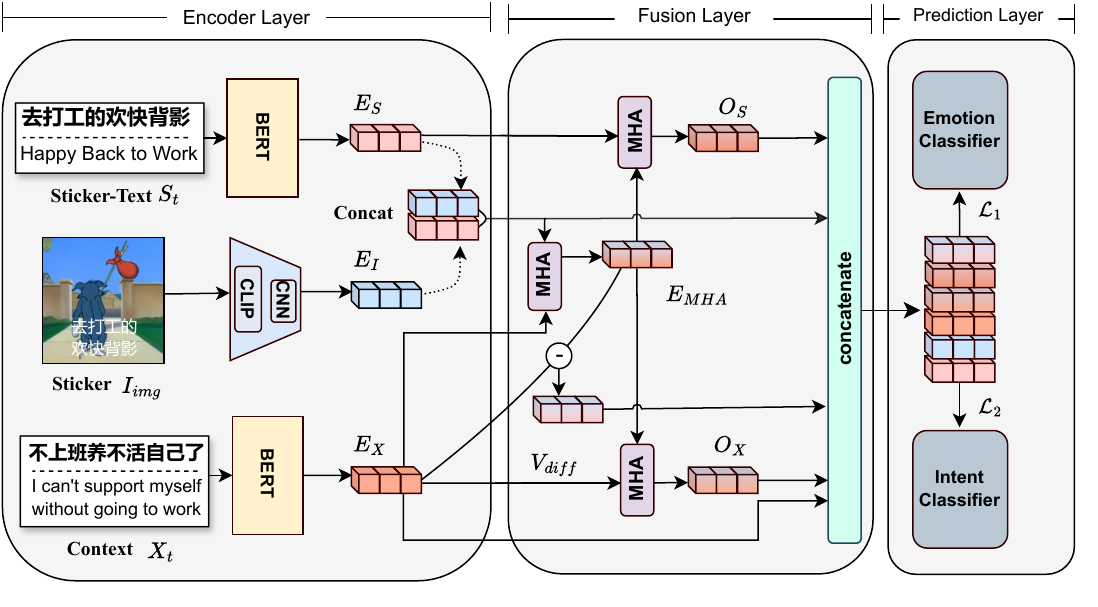} 
\caption{Abstract view of the multimodal baseline model MMSAIR.}
\label{fig.5}
\end{center}
\end{figure*}

\section{MMSAIR Model} 
As shown in Figure \ref{fig.5}, MMSAIR introduces a streamlined yet effective approach to multimodal sentiment analysis and intent recognition. The innovative use of a differential vector enhances the understanding of interactions between context and stickers, while a cascaded multi-head attention mechanism ensures robust feature fusion. This simplicity in design, combined with its effectiveness, allows MMSAIR to achieve high performance in analyzing the complex dynamics of social media communication.

\subsection{Task Description}

The objective of our baseline model MMSAIR is to jointly predict the multimodal sentiment label $y_{s}$ and intent label $y_{i}$ given the context $X_t = (x_1, x_2, \ldots, x_m)$, sticker image $I_{img}$, and sticker-text $S_t = (s_1, s_2, \ldots, s_n)$. Here, $x_i$ represents the $i$-th word in the context, and $s_i$ represents the $i$-th word in the sticker-text. The lengths of the context and sticker-text are denoted as $m$ and $n$, respectively.

\subsection{Encoding Layer}

\textbf{Text Encoder.} We utilize two separate BERT as the text encoder to capture contextual information. The sequence representation is derived from the [CLS] token of BERT's hidden layer, which is effective for downstream classification tasks.

The context encoding $E_X$ is computed as:
\begin{equation}
E_X = \text{BERT}(X_t)
\end{equation}

Similarly, the sticker-text encoding $E_S$ is computed as:
\begin{equation}
E_S = \text{BERT}(S_t)
\end{equation}

\noindent\textbf{Sticker Image Encoder.} To bridge textual and visual modalities, we use CLIP as the sticker image encoder. A CNN is applied for dimensionality reduction, resulting in $E_I$:

\begin{equation}
E_I = \text{Conv1d}(\text{CLIP}(I_{\text{img}}))
\end{equation}

\subsection{Representation Fusion Layer}

We fuse the sticker image and sticker-text representations by concatenating $E_I$ and $E_S$ to form $E_{I,S}$:

\begin{equation}
    E_{I,S} = \text{Concat}(E_I, E_S)
\end{equation}

Multi-head attention is applied with $E_X$ as the query and $E_{I,S}$ as the key and value, refining sticker features in the context:

\begin{equation}
O_{MHA} = \text{MultiHead}(E_{X}, E_{I,S}, E_{I,S})
\end{equation}

To capture the contrast between the context and the refined sticker features, we construct a differential vector $V_{diff}$ using learnable parameters $W_{diff}$ and $b_{diff}$:

\begin{equation}
V_{diff} = W_{diff} \cdot (O_{MHA} - E_X) + b_{diff}
\end{equation}

where $W_{diff}$ is a learnable weight matrix and $b_{diff}$ is a bias vector. Finally, another multi-head attention mechanism is applied to further refine the features, resulting in $O_S$ for sticker features and $O_X$ for context features:

\begin{equation}
O_{S} = \text{MultiHead}(E_{S}, E_{S}, O_{MHA})
\end{equation}
\begin{equation}
O_{X} = \text{MultiHead}(E_{X}, E_{X}, O_{MHA})
\end{equation}

\subsection{Prediction Layer}

The combined vector $E_{combined}$, formed by concatenating $E_{I,S}$, $E_X$, $V_{diff}$, $O_S$, and $O_X$, is passed through a fully connected neural network for classification:

\begin{equation}
E_{combined} = W_e \cdot \text{Concat}(E_{I,S}, E_X, V_{diff}, O_S, O_X) + b_e
\end{equation}

where $W_e$ and $b_e$ are learnable weights and biases. The sentiment and intent probability distributions are computed using softmax functions with learnable parameters:

\begin{equation}
P_{sentiment} = \text{Softmax}(W_s \cdot E_{combined} + b_s)
\end{equation}
\begin{equation}
P_{intent} = \text{Softmax}(W_i \cdot E_{combined} + b_i)
\end{equation}

where $W_s$, $b_s$, $W_i$, and $b_i$ are learnable weights and biases. The sentiment loss \(\mathcal{L}_1\) is computed using the predicted sentiment probabilities \(P_{sentiment}\) and the ground truth labels \(y_{s}\):

\begin{equation}
\mathcal{L}_1 = -\frac{1}{N} \sum_{i=1}^{N} y_{s}^{(i)} \log(P_{sentiment}^{(i)})
\end{equation}

\begin{table*}[!t]  \footnotesize
\centering   
\caption{\label{tb2}
Overall experimental results comparison.\textit{ Acc.} represents accuracy. \textit{F1} represents the weighted F1 score. The \textit{w/o} represents without. \textit{C\textunderscore{}F} represents context features. \textit{S\textunderscore{}F} represents sticker image features, and \textit{ST\textunderscore{}F} represents sticker-text features.
}

\begin{tabular}{l|l|cc|cc}   
\toprule
    & Models & Sentiment-Acc. & Sentiment-F1 & Intent-Acc. & Intent-F1\\
\midrule     & BERT & 65.05&64.48 &61.94 &61.84\\
          Context-only & ALBERT   & 60.55&60.00 &44.98 &44.48\\
          & RoBERTa     & 66.78 &66.55 &65.74 &65.40\\
\hline     & ViT &51.21 & 51.15& 15.57&16.09\\
           Image-only & Clip & 58.13 & 57.59& 21.11&20.11\\
          & ResNet50 & 42.56&38.97 & 12.80&11.03\\ 
\hline 
    \multirow{5}{*}{Multimodal}& mBERT& 68.86&\textbf{71.89} &65.05 &58.40 \\ 
    & EF-CAPTrBERT& 62.28&56.33 &49.48 &45.09 \\ 
    &PMF&65.95 &62.27 &52.60 &47.01 \\ 
    &CSMSA&63.96 &61.87 &57.48 &55.07 \\ 
    & MMSAIR(ours)&\textbf{70.58}&70.91&\textbf{72.31}&\textbf{72.29}\\ 
    \hline     & LLaVA &38.18 & 34.15& 14.55&16.49\\
   & LLaVA/fine-tuning&39.25& 36.71& 9.62& 10.10\\
         MLLM   & Yi-VL & 37.82 & 32.50& 5.10&6.09\\
       &  Yi-VL/fine-tuning&40.62&43.73& 7.48& 8.25\\
          & Qwen2.5-VL & 35.64&31.01 & 5.82&6.84\\ 
        &  Qwen2-VL/fine-tuning& 39.79& 47.76&7.61& 8.61\\
            & GPT-4o & 47.59&45.23 & 32.56&32.17\\ 
            & GPT-4o/one-shot & 53.48& 52.25& 38.62& 39.17\\ 
    \hline
      \multirow{5}{*}{Ablation Study}         &    w/o C\textunderscore{}F &67.39 &67.60&30.87 &30.07\\
        & w/o S\textunderscore{}F&68.78 &69.09 &68.90 &69.89\\
       &  w/o ST\textunderscore{}F& 60.55&60.46 &70.59 &70.57\\
        & w/o S\textunderscore{}F\&ST\textunderscore{}F (Context-only)& 65.40&64.83 & 67.47&67.24\\
       &  w/o C\textunderscore{}F\&ST\textunderscore{}F (Image-only)&59.17&58.16 & 27.36&26.78\\
\bottomrule
\end{tabular}   
\end{table*}

Similarly, the intent loss \(\mathcal{L}_2\) is calculated using the predicted intent probabilities \(P_{intent}\) and the ground truth labels \(y_{i}\):

\begin{equation}
\mathcal{L}_2 = -\frac{1}{N} \sum_{i=1}^{N} y_{i}^{(i)} \log(P_{intent}^{(i)})
\end{equation}

where \(N\) is the number of samples in the batch. The overall loss function \(\mathcal{L}\) is defined as a weighted sum of the sentiment loss \(\mathcal{L}_1\) and the intent loss \(\mathcal{L}_2\):

\begin{equation}
\mathcal{L} = \alpha \mathcal{L}_1 + \beta \mathcal{L}_2
\end{equation}

where $\alpha$ and $\beta$ control the weight of each task.

\begin{table}[!t] \footnotesize
\begin{center}
\caption{\label{tb3}
Experimental results for Individual Sentiment Analysis, Intent Recognition Tasks, and the Joint Task of Both.}
\begin{tabular}{l|cccc|cccc}
\toprule
Model & \multicolumn{4}{c|}{MMSAIR} & \multicolumn{4}{c}{GPT-4o} \\
\hline
\multirow{2}{*}{Task}&\multicolumn{2}{c}{Sentiment} & \multicolumn{2}{c|}{Intent}&\multicolumn{2}{c}{Sentiment} & \multicolumn{2}{c}{Intent}\\
 & Acc.&F1&Acc.&F1& Acc.&F1&Acc.&F1\\
\hline
SA &68.23 &67.87 & -&-&44.98 &44.69 & -&-\\
\hline
IR & -& -&68.95 &68.52 &- &- & 28.71&26.94\\
\hline
MSAIRS&\textbf{70.58}&\textbf{70.91}&\textbf{72.31}&\textbf{72.29}& \textbf{47.59}&\textbf{45.23} & \textbf{32.56}&\textbf{32.17}\\
\bottomrule
\end{tabular}
 \end{center}
\end{table}

\section{Experiment}

\subsection{Experimental Setups}

We compare our model with several popular unimodal and multimodal models. We use \textbf{BERT} \cite{2018BERT}, \textbf{ALBERT} \cite{2019ALBERT}, and \textbf{RoBERTa} \cite{2019RoBERTa} as context-only baselines, using the context as input. For the image-only models, we utilize \textbf{ViT} \cite{2020An}, \textbf{CLIP} \cite{2021Learning}, and \textbf{ResNet50} \cite{he2016deep} as baselines, taking stickers as input.
The same linear layer and classifier are added to obtain classification results. We choose \textbf{mBERT} \cite{yu2019adapting}, \textbf{EF-CAPTrBERT} \cite{khan2021exploiting}, \textbf{PMF} \cite{li2023efficient} and \textbf{CSMSA} \cite{ge-etal-2022-towards} for multimodal comparison. We select \textbf{LLaVA-7b} \cite{liu2023visualinstructiontuning}, \textbf{Yi-VL-6b} \cite{young2024yi}, \textbf{Qwen2.5-VL-7b} \cite{wang2024qwen2} and \textbf{GPT-4o} for MLLM comparison.

All models are trained for 50 epochs with the Adam \cite{kingma2014adam} optimizer, a learning rate of 1e-5, a train batchsize of 16, and a valid batchsize of 2. We set the $\alpha$ and $\beta$ in MMSAIR both to 1, indicating that sentiment analysis and intent recognition equals. For MLLMs, we conduct both zero-shot and fine-tuning experiments (one-shot for GPT-4o) using the following prompt: 

\noindent\textbf{"Determine the multimodal sentiment and intent expressed in this social media chat combined with the corresponding sticker. The sentiment should be one of: positive, negative, or neutral. The intent should be selected from \textit{intent\_set}. The text is: \textit{context}, and the sticker is \textit{<img>}. Output the sentiment and intent directly, separated by a comma." }

Here, \textit{intent\_set} refers to all intents listed in Table \ref{tbintent}. \textit{context} and \textit{<img>} represent the chat context and sticker image.

We include the unimodal sentiment analysis experiments in Supplementary Materials, demonstrating the  applicability of MMSAIR.

\subsection{Overall Results}
\label{exre}

\begin{figure*}[!ht]
\begin{center}
\includegraphics[scale=0.5]{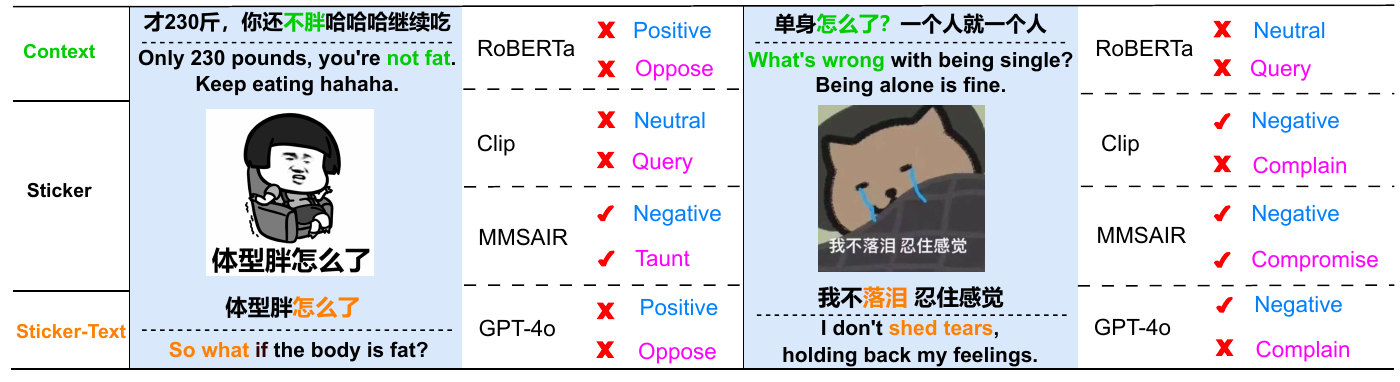} 
\caption{Experimental results of typical examples in our dataset using different models.}
\label{case}
\end{center}
\end{figure*}

Table~\ref{tb2} shows the experimental results on MSAIRS. Context-only models perform well due to the direct nature of text, despite potentially producing one-sided results. In contrast, image-only models struggle with abstract sticker content, particularly in intent recognition tasks, highlighting the importance of textual data and effective sticker-text processing. Multimodal models generally outperform unimodal approaches, with MMSAIR excelling in both sentiment and intent tasks through effective text-sticker fusion. While mBERT shows comparable sentiment analysis performance, it underperforms in intent recognition. MLLMs consistently perform poorly across both tasks, with even the advanced GPT-4o with one-shot prompt achieving only 53.48\% sentiment accuracy and 38.62\% intent accuracy, less than half of MMSAIR's intent recognition performance. LLaVA even shows performance degradation with fine-tuning. This gap likely stems from MLLMs' limited exposure to social media stickers during pre-training, particularly those conveying complex emotions through irony, sarcasm, and cultural references, as well as their struggle with inherent ambiguity when stickers contradict textual sentiment. These results establish MMSAIR as a simple yet highly effective baseline. We also conduct significance tests in supplementary materials.

\subsection{Multimodal Impact}

Table~\ref{tb2} shows that removing context features severely impacts intent recognition, confirming context provides crucial intent information. Removing image or sticker-text features causes smaller performance decreases, demonstrating both components enhance prediction quality. Without sticker-related features (S\_F and ST\_F), performance decreases across both tasks, indicating stickers provide valuable complementary information to text. With only image features, sentiment analysis performs moderately but intent recognition suffers dramatically, confirming images alone cannot effectively determine intent. These findings establish that the context, sticker images, and sticker-text, are all essential for optimal MSAIRS performance. MMSAIR demonstrates strong robustness by effectively handling both multimodal and unimodal inputs.

\subsection{Subtask Influence}

Sentiment analysis and intent recognition tasks demonstrate significant mutual influence. As shown in Table ~\ref{tb3}, when performed independently, both tasks yield poorer results compared to our joint approach. For MMSAIR, joint modeling improves sentiment accuracy by 2.35\% and intent accuracy by 3.36\%. Similarly, GPT-4o shows modest improvements in both sentiment and intent recognition when tasks are performed jointly. These results clearly demonstrate that sentiment determination significantly impacts intent recognition and vice versa. The consistent improvements across both traditional and MLLMs underscore the value of our joint task approach, proving that MSAIRS effectively captures the inherent interdependence between sentiment and intent in social media communications.

\section{Case Study}

In Figure ~\ref{case}, we present two typical examples from MSAIRS and corresponding experimental results of the best context-only model (RoBERTa), image-only model (Clip), MMSAIR model, and GPT-4o. In the figure, the red checkmarks indicate predictions that match the Ground Truth, and the red crosses indicate wrong results.

In the first example, the context-only RoBERTa model focuses on phrases like "you're not fat" and "hahaha," incorrectly predicting a positive sentiment with an opposing intent. The image-only Clip model, processing only the sticker with its indifferent character and "so what?" text, predicts neutral sentiment and querying intent. GPT-4o incorrectly interprets the interaction as expressing positive sentiment with opposing intent, similar to RoBERTa. Only MMSAIR correctly recognizes that although the text says "not fat," the "hahaha" combined with the sticker's ironic tone reveals a negative, taunting intent, successfully capturing the mockery in the message.
In the second example, RoBERTa interprets the question in the context as a neutral query. Clip, seeing only a crying cartoon character with "shed tears" text, predicts negative complaint. GPT-4o also misinterprets the communication as expressing negative sentiment with a complaining intent. However, MMSAIR correctly identifies that the speaker is expressing helplessness about being single, using the crying sticker to convey negative sentiment with a compromising intent. By effectively combining context and visual cues, MMSAIR produces the only accurate assessment matching the Ground Truth.

These examples demonstrate how existing models struggle to predict sentiment and intent in social media communications with stickers. The complexities of irony, cultural context, and multimodal interactions pose significant challenges. MMSAIR's ability to handle these nuanced examples highlights its effectiveness as a multimodal baseline for analyzing sticker-based interactions.

\section{Conclusion}

In this paper, we investigate the impact of stickers on sentiment analysis and intent recognition in social media communications. We introduce MSAIRS, a novel task and manually annotated dataset, alongside an effective multimodal baseline model MMSAIR. Our experiments reveal that contextual and visual information must be integrated for accurate analysis. Notably, even advanced MLLMs like GPT-4o struggled with this task, highlighting the unique challenges of interpreting social media stickers. The improved performance when jointly modeling sentiment and intent confirms their interdependence in real-world communications. Future research could explore more sophisticated fusion techniques between modalities and investigate additional contextual features to further enhance performance in this challenging multimodal understanding task.

\section*{Limitations and Ethical Considerations}
The chat records and stickers are exclusively sourced from Chinese social media platforms. Given the cultural differences between Chinese and Western contexts, certain expressions and interpretations of sentiment and intent may not be directly transferable across languages. As a result, the dataset primarily contributes to the study in Chinese social media interactions. To ensure ethical integrity, all data and stickers have undergone rigorous manual review to eliminate content that could cause physiological or psychological discomfort. The dataset strictly excludes any material related to violence, pornography, or politically sensitive topics, ensuring compliance with ethical guidelines and platform regulations.

\begin{acks}
This work was supported by the Project 62276178 under the National Natural Science Foundation of China, the Key Project 23KJA520012 under the Natural Science Foundation of Jiangsu Higher Education Institutions, the project 22YJCZH091 of Humanities and Social Science Fund of Ministry of Education and the Priority Academic Program Development of Jiangsu Higher Education Institutions.
\end{acks}

\bibliographystyle{ACM-Reference-Format}
\bibliography{latex/Sticker}

\end{document}